\documentclass{article}
\usepackage{spconf,amsmath,graphicx}
\usepackage{amsfonts, amssymb}
\usepackage{spconf,amsmath,graphicx, mathrsfs}
\usepackage{algorithmic}
\usepackage{algorithm}
\usepackage{bm}
\usepackage{multirow}      
\usepackage{multicol}
\usepackage{booktabs}
\usepackage{subfig}
\usepackage{graphicx}
\usepackage{caption}
\usepackage{hyperref}

\newcommand{\etal}{\textit{et al.}}

\title{AdaPlus: Integrating Nesterov Momentum and Precise Stepsize Adjustment on AdamW Basis}
\name{Lei Guan\thanks{This work was supported in part by the State Administration of Science, Technology and Industry for National Defence (No. WDZC20235250118).}}
\address{Department of Mathematics, National University of Defense Technology 
	\\
	guanleimath@163.com
}
\begin{document}
%
\maketitle
\begin{abstract}
This paper proposes an efficient optimizer called AdaPlus which integrates Nesterov momentum and precise stepsize adjustment on AdamW basis. AdaPlus combines the advantages of AdamW, Nadam, and AdaBelief and, in particular, does not introduce any extra hyper-parameters.  We perform extensive experimental evaluations on three machine learning tasks to validate the effectiveness of AdaPlus. The experiment results validate that AdaPlus (i) among all the evaluated adaptive methods, performs most comparable with (even slightly better than) SGD with momentum on image classification tasks and (ii) outperforms other state-of-the-art optimizers on language modeling tasks and illustrates pretty high stability when training GANs. The experiment code of AdaPlus will be accessible at: \url{https://github.com/guanleics/AdaPlus}.


\end{abstract}
\begin{keywords}
deep learning, adaptive method, Nesterov momentum,  generalization, stability
\end{keywords}
\section{Introduction}
\label{sec:intro}

First-order gradient methods have been broadly used in the training of deep neural networks. The popular first-order gradient methods, in general, can be categorized as accelerated schemes (e.g. stochastic gradient descent with momentum (SGDM)~\cite{sutskever2013importance}) and adaptive methods (e.g. Adam~\cite{kingma2014adam} and AdamW~\cite{loshchilov2017decoupled}). Adaptive methods generally compute an individual stepsize (a.k.a. learning rate) for each parameter and play a significantly important role in the training of modern deep neural networks. Especially, Adam~\cite{kingma2014adam} can attain rapid training speed and has been acting as the default choice for deep learning training. 

Much progress on adaptive methods is built upon Adam. For instance, considering the fact that Adam does not generalize as well as SGD with momentum when handling image classification tasks, Loshchilov \etal~\cite{loshchilov2017decoupled} propose the AdamW optimizer which introduces decoupled weight decay into Adam and achieves competitive performance as SGDM when tackling image classification tasks. Based on the observation that Nesterov's accelerated gradient (NAG)~\cite{nesterov27method} is empirically superior to the regular momentum, Timothy Dozat~\cite{dozat2016incorporating} incorporates Nesterov momentum into Adam and proposes the Nadam optimizer. To achieve fast convergence, comparable accuracy to SGD, and provide high stability in the training of a GAN, Zhuang \etal~\cite{zhuang2020adabelief}  propose the AdaBelief optimizer. AdaBelief views the exponential moving average (EMA) of the noisy gradient as the prediction of the gradient in the next time step and adapts the stepsize according to the ``belief'' in the current gradient direction. The advantage of AdaBelief over Adam mainly lies in the `` large gradient, small curvature'' case where Adabelief, unlike Adam, increases the stepsize as the ideal optimizer does.

It's obvious that AdamW, Nadam, and AdaBlief all build based on Adam but enjoy different advantages in terms of boosting adaptive methods. To combine the benefits of these three adaptive methods, we propose a new optimizer AdaPlus which, on the AdamW basis, simultaneously integrates Nesterov momentum as in Nadam and precise stepsize adjustment as in AdaBelief. To validate the effectiveness of AdaPlus, we experiment with three typical machine learning tasks, including image classification with CNNs on CIFAR10, language modeling with LSTM on Penn TreeBank, and generative adversarial networks (GAN) on CIFAR10. We compare AdaPlus with eight state-of-the-art optimzers including SGDM~\cite{sutskever2013importance}, Adam~\cite{kingma2014adam}, Nadam~\cite{dozat2016incorporating}, RAdam~\cite{liu2019variance}, AdamW~\cite{loshchilov2017decoupled}, AdaBelief~\cite{zhuang2020adabelief}, AdamW-Win~\cite{zhou2022win}, and Lion~\cite{chen2023symbolic}. The experiment results demonstrate that AdaPlus outperforms the other optimizers in simultaneously achieving the goal of (i) fast convergence, (ii) good generalization ability, and (iii) high stability in the training of GANs. For example, on the image classification task, AdaPlus yields an average test accuracy improvement of 1.97\% (up to 2.36\%), 1.85\% (up to 2.0\%), and 0.52\% (up to 0.89\%) over AdamW, Nadam, and AdaBelief, respectively. Furthermore, on the GAN training, AdaPlus always attains a low FID score, illustrating pretty good stability. 

The contributions of this paper can be summarized as follows:
\begin{itemize} 
	\item [(1)] We propose a new adaptive optimizer named AdaPlus, which builds based on the AdamW optimizer and further incorporates Nesterov momentum as in Nadam and precise stepsize adjustment as in AdaBelief. AdaPlus is able to combine the advantages of AdamW, Nadam, and AdaBelief. To the best of our knowledge, this is the first adaptive method that simultaneously combines the advantages of decoupled weight decay, Nesterov momentum, and precise stepsize adjustment. 
	\item [(2)] We conducted extensive experimental evaluations on three
 machine-learning tasks to validate the effectiveness of AdaPlus. AdaPlus, among all evaluated optimizers, is the best adaptive method that performs most comparable with SGDM and performs the best in simultaneously achieving the goal of fast convergence, good generalization ability, and high stability.
\end{itemize}

\section{Methods}
\label{sec:format}

\textbf{Notations}
In this paper, we let $f(\bm \theta)\in \mathbb{R}^d$ be the loss function to minimize where $\bm \theta$ ($\bm \theta \in \mathbb{R}$) is the parameter to learn. 
We let $\mathbf g_t$ denote the gradient at step $t$ and $\mathbf m_t$  refer to the EMA of $\mathbf g_t$. The learning rate is represented by $a$, the weight decay is denoted by $u$, and $\epsilon$ is the smoothing term. Moreover, $\mathbf v_t$ and $\mathbf s_t$ respectively denote the EMA of $\mathbf g_t^2$ and $(\mathbf g_t-\mathbf m_t)^2$. $\beta_1$ and $\beta_2$ are the smooting parameters which are typically set to $\beta_1=0.9$ and $\beta_2 = 0.999$.

\subsection{Modifying AdamW's Momentum}\label{subsec:nesterov}
Inspired by~\cite{dozat2016incorporating}, we first rewrite NAG as
\begin{equation}
	\begin{aligned}
			&{\mathbf g}_t \leftarrow \nabla_{\bm \theta-1}f({\bm \theta}_{t-1}), \\
			&{\mathbf m}_{t} \leftarrow u  \mathbf m_{t-1} + a \mathbf g_t,\\
			&\bm \theta_t \leftarrow \bm\theta_{t-1} - (u \mathbf m_t + a\mathbf g_t). \\
		\end{aligned}
		\label{eq:nesterov}
	\end{equation}
	Equation~\eqref{eq:nesterov} reveals that NAG updates the parameter with $u\mathbf m_t$ rather than $u\mathbf m_{t-1}$ used in the classical momentum. 
	
	
	To incorporate Nesterov momentum into AdamW, we replace the classical momentum $\mathbf m_t$ in AdamW with Nesterov momentum $\beta_1 \mathbf m_{t-1} +(1-\beta_1)\mathbf g_t$. Then we rewrite AdamW's update step in terms of $\mathbf m_{t-1}$ and $\mathbf g_t$, which is
	\begin{equation}
		{\bm \theta_t} \leftarrow {\bm \theta_{t-1}} -\frac{a}{\sqrt{\hat{\mathbf v}_t} + \epsilon} \big( \frac{\beta_1 \mathbf m_{t-1}}{1-\beta_1^t} + \frac{(1-\beta_1)\mathbf g_t}{1-\beta_1^t} \big).
	\end{equation}

After substituting the next momentum step for the current one, we have
\begin{equation}
	{\bm \theta_t} \leftarrow {\bm \theta_{t-1}} -\frac{a}{\sqrt{\hat{\mathbf v}_t} + \epsilon} \big( \frac{\beta_1 \mathbf m_{t}}{1-\beta_1^t} + \frac{(1-\beta_1)\mathbf g_t}{1-\beta_1^t} \big).
\end{equation}

That can be equivalently rewritten as
\begin{equation}
	\begin{aligned}
		&\bar{\mathbf m}_t \leftarrow \beta_1 {\mathbf m_t} + (1-\beta_1)\mathbf g_t, \\
		& \hat{\mathbf m}_t \leftarrow \frac{ \bar{\mathbf m}_t}{1-\beta_1^t}, \\
		&{\bm \theta_t} \leftarrow {\bm \theta_{t-1}} -\frac{a\hat{\mathbf m}_t}{\sqrt{\hat{\mathbf v}_t} + \epsilon}.
	\end{aligned}
\end{equation}

\subsection{Precise Stepsize Adjustment}
On the basis of Section~\ref{subsec:nesterov},  we further integrate the stepsize adjusting mechanism proposed in~\cite{zhuang2020adabelief} and finally propose a new optimizer named AdaPlus. Algorithm~\ref{algorithm:adaplus} summarizes the details of AdaPlus.  It's worth noting that no extra hyper-parameters are introduced in AdaPlus in comparison with AdamW and AdaBelief. As shown in Line 7 of Algorithm~\ref{algorithm:adaplus}, AdaPlus regards $\mathbf m_t$ as the forecast for $\mathbf g_t$, escalates the stepsize when $\mathbf g_t$ approaches $\mathbf m_t$ and decreases the stepsize when $\mathbf g_t$ deviates from the prediction $\mathbf m_t$.

\begin{algorithm}[H]
	\caption{The AdaPlus Optimizer} 
	\label{algorithm:adaplus} 
	\begin{algorithmic}[1]
		\REQUIRE {\text{initial learning rate} $a=0.001$, $\beta_1=0.9$, $\beta_2=0.999$, $\epsilon=10^{-8}$, \text{weight decay factor} $\lambda \in \mathbb{R}$}
		\STATE \textbf{Initialize} time step $t\leftarrow 0$,  $\bm \theta_0$, $\mathbf m_0 \leftarrow 0$, $\mathbf v_0 \leftarrow 0$, $t\leftarrow 0$.
		\WHILE {$\theta_t$ not converged}
		\STATE{$t \leftarrow t+1$}
		\STATE{ $\mathbf g_t\leftarrow \nabla_\theta f_t(\bm \theta_{t-1})$} 
		\STATE{$\bm \theta_t \leftarrow \bm \theta_{t-1} - \gamma\lambda \bm \theta_{t-1}$}
		\STATE {$\mathbf m_t \leftarrow \beta_1 \mathbf m_{t-1} + (1-\beta_1)\mathbf g_t$}
		\STATE {$\mathbf s_t \leftarrow \beta_2 \mathbf s_{t-1} + (1-\beta_2) (\mathbf g_t - \mathbf m_t)^2 + \epsilon$}
		\STATE{$\bar{\mathbf m}_t \leftarrow \beta_1 \mathbf m_t + (1-\beta_1) \mathbf g_t$}
		\STATE {$\hat{\mathbf m_t} \leftarrow \frac{\bar{\mathbf m}_t}{1-\beta_1^t}$, $\hat{\mathbf s_t} \leftarrow \frac{\mathbf s_t}{1-\beta_2^t}$ }
		\STATE {$\bm \theta_t \leftarrow \bm \theta_{t-1} - \frac{a\hat{\mathbf m_t}}{\sqrt{\hat{\mathbf s_t}}+\epsilon}$ }
		\ENDWHILE
	\end{algorithmic}
\end{algorithm}

\begin{figure}[ht]
	\centering
	\includegraphics[scale=0.25]{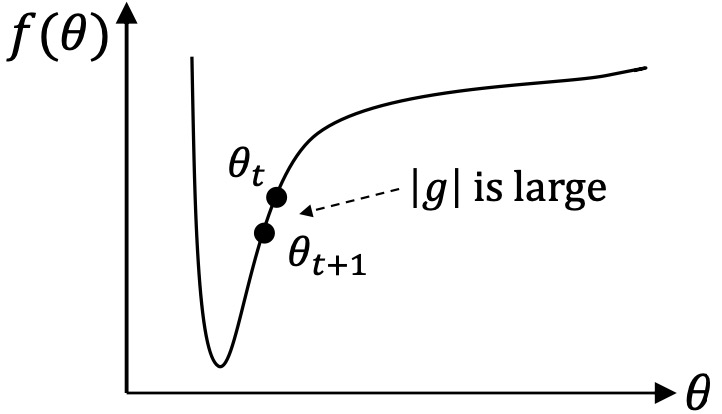}
	\caption{Illustration of ``large gradient, small curvature'' case where current stepsize is small and $|g(\bm \theta_t) - g(\bm \theta_{t+1})|$ is small. An ideal optimizer should increase the stepsize.}
	\label{fig:large_grad}
\end{figure}

\noindent\textbf{Comparison with AdamW, Nadam, and AdaBelief}. \ We mainly consider the ``large gradient, small curvature'' case in which AdaBelief~\cite{zhuang2020adabelief}, with precise stepsize adjustment, performs differently from other adaptive methods (e.g. Adam). The details are shown in Figure~\ref{fig:large_grad}, 

\begin{table*}[!h]
	\centering
	\caption{Maximum test accuracy on CIFAR-10. \textbf{Higher} is better.}
	\label{table:cifar-acc}
	\setlength{\tabcolsep}{2.4mm}
	\begin{tabular}{c|ccccccccc}
		\toprule
		Models & AdaPlus &SGDM &  Adam & Nadam & AdamW &  RAdam   & AdaBelief & Lion& AdamW-Win    \\
		\midrule
		VGG-11  & \textbf{90.55}\% & 90.48\% &  88.89\% &88.19\% &   88.64\% & 90.05\%&  90.07\% & 87.71\% &89.72\% \\
		ResNet-34 & \textbf{94.99}\%  & 94.96\% & 92.99\%  &  93.19\% &94.50\%  & 93.33\% &  94.10\% & 94.10\% & 94.72\% \\
		DenseNet-121 &94.91\%  & \textbf{95.37}\%  & 93.02\%  & 93.17\% & 94.11\%  & 93.70\% & 94.71\% &94.54\% & 94.75\%\\
		\bottomrule
	\end{tabular}
\end{table*}

\begin{figure*}[!h]
	\centering
	\subfloat[VGG-11  on CIFAR-10]{\includegraphics[width=.3\textwidth]{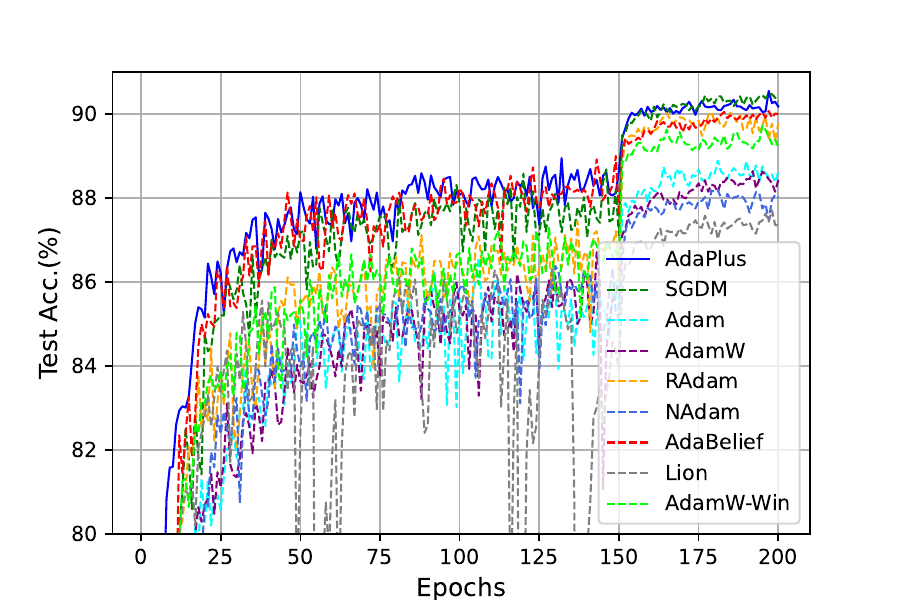}\label{comp-vgg11-acc}}
	\subfloat[ResNet-34 on CIFAR-10]{\includegraphics[width=.3\textwidth]{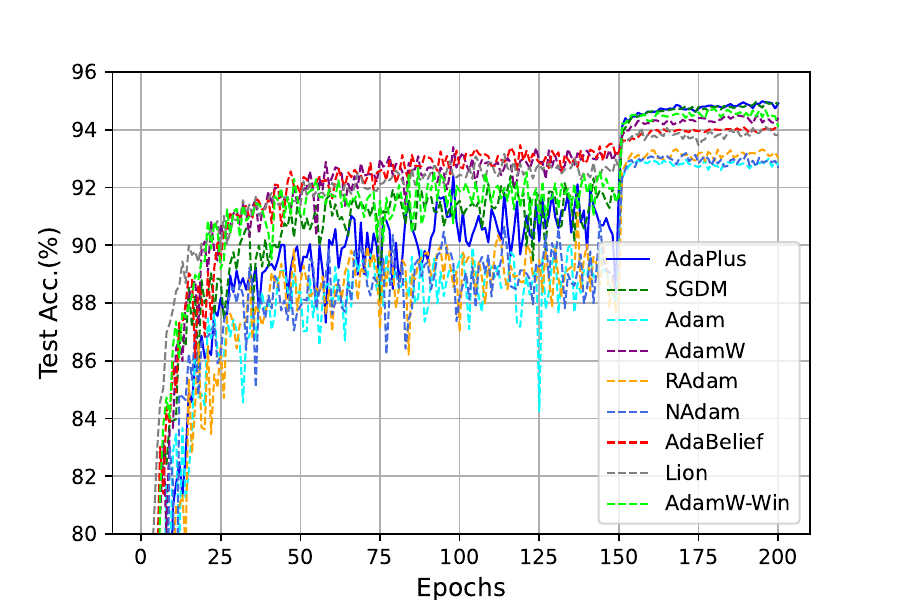}\label{comp-resnet34-acc}} 
	\subfloat[DenseNet-121 on CIFAR-10]{\includegraphics[width=.3\textwidth]{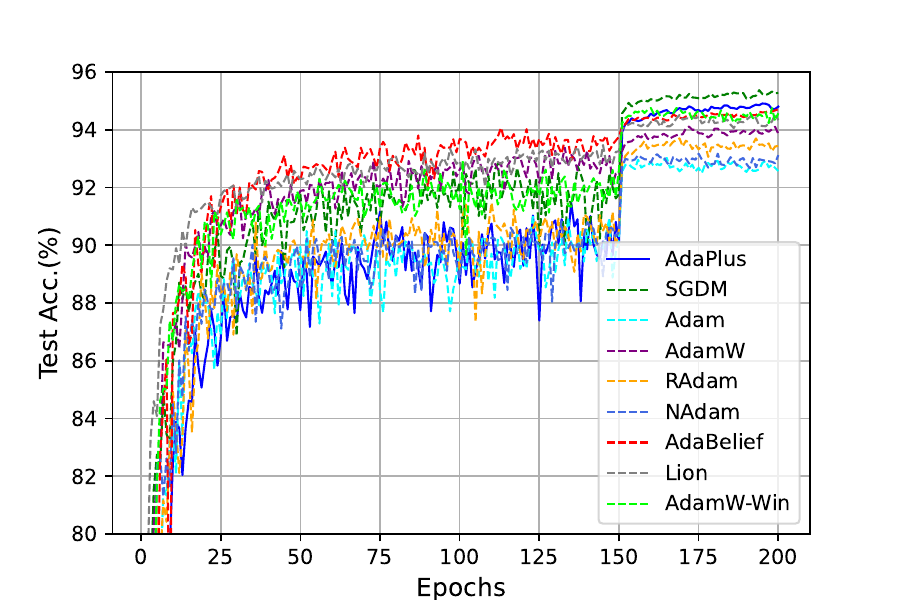}\label{comp-densenet-acc}}
	\caption{Validation accuracy vs. epochs of training VGG-11, ResNet-34, and DenseNet-121  on CIFAR-10.}
	\label{comp-acc-cifar}
\end{figure*}

We note that the update formulas for AdamW, Nadam, AdaBelief, and AdaPlus are:
\begin{equation}
	\begin{aligned}
		&\Delta \bm\theta_t^{\text{AdamW}, \ \text{Nadam}}=-\frac{a\hat{\mathbf m}_t}{\sqrt{\hat{\mathbf v}_t} + \epsilon}, \\
		&\Delta \bm\theta_t^{\text{AdaBelief}, \ \text{AdaPlus}}=-\frac{a\hat{\mathbf m}_t}{\sqrt{\hat{\mathbf s}_t} + \epsilon}
	\end{aligned}
   \label{eq:update}
\end{equation}

Equation~\eqref{eq:update} reveals that the update directions in AdamW and Nadam are $\mathbf m_t / (\sqrt{\mathbf v_t} + \epsilon)$, where $\mathbf v_t$ is the EMA of $\mathbf g_t^2$;  the update direction in AdaPlus is $\mathbf m_t / (\sqrt{\mathbf s_t} + \epsilon)$, where $\mathbf s_t$ is the EMA of $(\mathbf g_t - \mathbf m_t)^2$. For the ``large gradient, small curvature''  case, $|\mathbf g_t|$ and $\mathbf v_t$ are large, but $|\mathbf g_t - \mathbf g_{t-1}|$ and $\mathbf s_t$ are small. In this case, an ideal optimizer should increase its stepsize. It's clear that AdamW takes a smaller stepsize as $\mathbf v_t$ is large. In contrast, as done in an ideal optimizer, AdaPlus and AdaBelief tend to increase its stepsize as $\mathbf s_t$ is small. This demonstrates that AdaPlus can take precise stepsize as AdaBelief does.

\section{Experiments}
\label{sec:typestyle}

We perform extensive comparisons with eight state-of-the-art optimizers: SGDM~\cite{sutskever2013importance}, Adam~\cite{kingma2014adam}, Nadam~\cite{dozat2016incorporating},  AdamW~\cite{loshchilov2017decoupled}, RAdam~\cite{liu2019variance}, AdaBelief~\cite{zhuang2020adabelief}, AdamW-Win~\cite{zhou2022win}, and Lion~\cite{chen2023symbolic}. The experimental evaluations include three machine learning tasks, (a) image classification on CIFAR-10 with VGG~\cite{simonyan2014very}, ResNet~\cite{he2016deep}, and DenseNet~\cite{huang2017densely}, (b) language modeling on Penn TreeBank with LSTM~\cite{ma2015long} models, and (c) Wasserstein-GAN (WGAN)~\cite{arjovsky2017wasserstein} and the improved version with gradient penalty (WGAN-GP)~\cite{salimans2016improved} on CIFAR-10 dataset.

We implement AdaPlus in PyTorch on the AdamW basis.  The experimental evaluations follow that reported in~\cite{zhuang2020adabelief}. On the image classification task, we train all CNN models for 200 epochs with a mini-batch size of 128 and decay the learning rate by 0.1 at the 150th epoch. For the language modeling task, we train LSTMs with 1, 2, and 3 layers on Penn TreeBank dataset where in each experiment, the LSTM models are trained for 200 epochs with a batch size of 20, and the learning rate is decayed by 0.1 at the 100th and 145th epoch.

We note that SGDM, Adam, RAdam, and AdaBelief use the same hyper-parameter tunning strategy as reported~\cite{zhuang2020adabelief} which we do not report in detail due to space limit. Nadam and AdamW-Win set their default parameter values in the literature. On the image classification task, we set $\beta_1=0.9$ and $\beta_2=0.999$. For Lion, we use the suggested parameter in~\cite{chen2023symbolic} for image classification and language modeling tasks and search for optimal $\beta_1$ among \{0.5, 0.6, 0.7, 0.8, 0.9\}. For AdaPlus, we set weight decay as $1e-2$ and set $\epsilon$ as $1e-8$. We initialize the learning rate with 0.001 for VGG-16 and 0.01 for ResNet-34 and DenseNet-121. On the language modeling task, the hyper-parameters for AdaPlus are $\beta_1=0.9$, $\beta_2=0.999$, $\epsilon = 1e-16$. We initialize the learning rate with $1e-3$ and set the weight decay to $1e-2$. For the training of GANs, we seek optimal $\beta_1$ among \{0.5, 0.6, 0.7, 0.8, 0.9\} and set $a=2e-4$, $\beta_2=0.999$, $\epsilon=1e-12$, and $\lambda=1e-2$.


\subsection{Experiments for Image Classification}
\label{ssec:subhead}


%
%
%

Table~\ref{table:cifar-acc} summarizes the experiment on the CIFAR-10 dataset. Figure~\ref{comp-acc-cifar} depicts the learning curves of test accuracy vs. epochs for training CNN models of each evaluated optimizer. When training VGG-11 and ResNet-34, AdaPlus always attains higher test accuracy than the other optimizers. In addition, when training DenseNet-121, AdaPlus performs the best among all adaptive methods.  In particular,  AdaPlus achieves an average of 1.85\% (up to 2.0\%), 1.97\% (up to 2.36\%), 1.07\% (up to 1.91\%),  1.12\% (up to 1.66\%), 0.52\% (up to 0.89\%), 1.31\% (up to 2.68\%), and 0.42\% (up to 0.83\%) accuracy improvement over Adam,  Nadam, AdamW, RAdam, AdaBelief, Lion, and AdamW-Win, respectively. 


\begin{table*}[!h]
	\centering
	\caption{Minimum perplexity on Penn TreeBank. \textbf{Lower} is better.}
	\label{table:lstm-ppl}
	\setlength{\tabcolsep}{3.2mm}
	\begin{tabular}{c|ccccccccc}
		\toprule
		LSTM &  AdaPlus  & SGDM &   Adam   & Nadam &  AdamW  & RAdam &   AdaBelief & Lion &AdamW-Win \\
		\midrule
		1 layer & \textbf{86.22} &  104.13 &88.54   &87.98&  88.49&  88.56&  88.59 & 89.64 & 86.73\\
		2 layers & \textbf{71.72} & 83.80 & 73.72  &  73.91&  73.43 & 74.20& 72.97 & 73.65& 71.93\\
		3 layers & 68.08 & 86.93 & 70.24  & 69.82& 69.67 &  70.01  & 69.10 & 69.77& \textbf{68.03}\\
		\bottomrule
	\end{tabular}
\end{table*}

\begin{figure*}[!h]
	\centering
	\subfloat[1-layer LSTM]{\includegraphics[width=.3\textwidth]{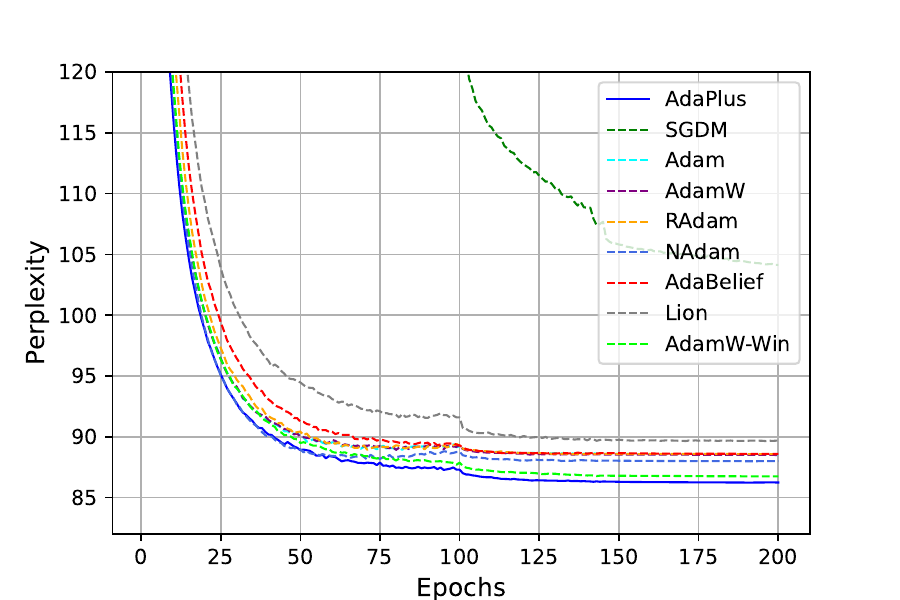}\label{comp-lstm1-ppl}}
	\subfloat[2-layer LSTM]{\includegraphics[width=.3\textwidth]{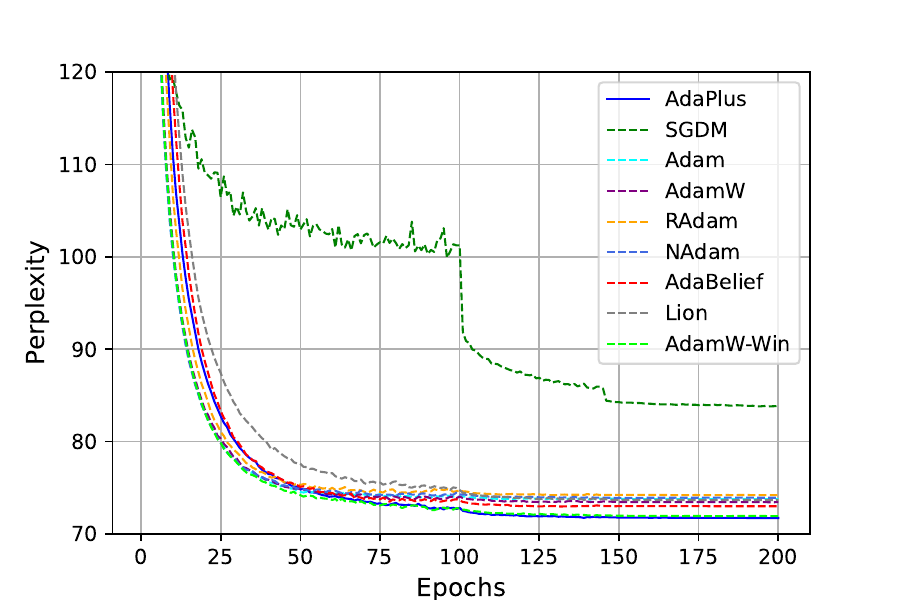}\label{comp-lstm2-ppl}} 
	\subfloat[3-layer LSTM]{\includegraphics[width=.3\textwidth]{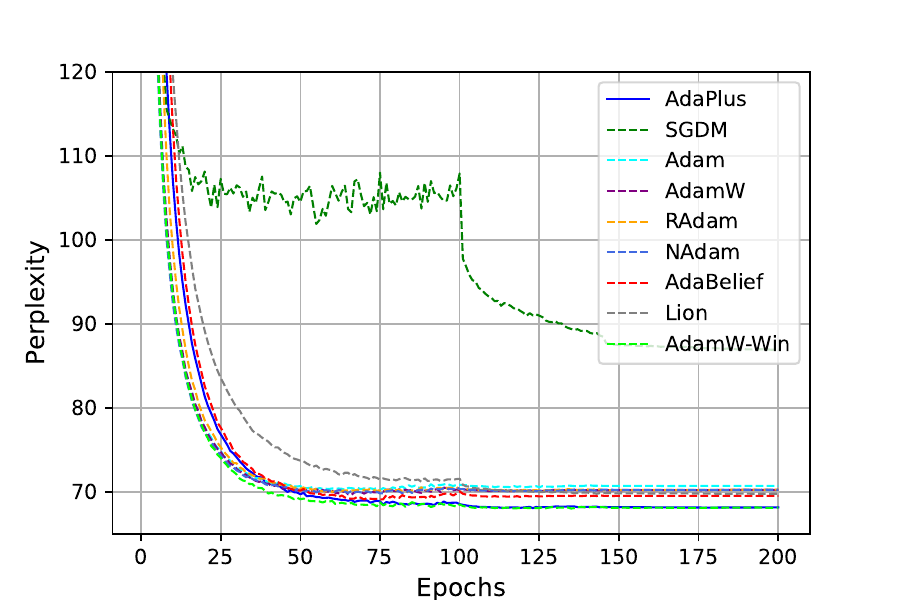}\label{comp-lstm3-ppl}}
	\caption{Perplexity vs. epochs of training LSTM on Penn TreeBank.}
	\label{comp-lstm-ppl}
\end{figure*}

\begin{table*}[!h]
	\centering
	\caption{FID (lower is better) of WGAN and WGAN-GP on CIFAR-10.}
	\label{table:gan-fid}
	\setlength{\tabcolsep}{3.2mm}
	\begin{tabular}{c|ccccccccc}
		\toprule
		Model &  AdaPlus  & SGDM &   Adam  & Nadam &  AdamW  &  RAdam & AdaBelief &Lion & AdamW-Win  \\
		\midrule
		WGAN & 82.96 & 299.88 & 94.15 &95.17 &  93.72 &  108.09 &  86.92 & 77.48 & \textbf{60.10}\\
		WGAN-GP &  \textbf{63.70} &  257.67 &  76.60   & 76.54&  68.85 &  94.29 & 66.63 & 249.58& 64.40\\
		\bottomrule
	\end{tabular}
\end{table*}

\subsection{Experiments for Language Modeling}
\label{ssec:computation}
Figure~\ref{comp-lstm-ppl} depicts the learning curves about perplexity vs. epochs. Table~\ref{table:lstm-ppl} presents the obtained minimum perplexity (lower is better). 
The experimental results shown in Table~\ref{table:lstm-ppl} again validate the generalization ability of AdaPlus. When training the 1-layer and 2-layer LSTM models, AdaPlus consistently attains the lowest perplexity among all evaluated optimizers. For training 3-layer LSTM, AdaPlus ranks second with very comparable low perplexity as AdamW-Win.



\subsection{Experiments for GANs on CIFAR-10}
\label{ssec:gan}
In this section, we experiment with the Wasserstein-GAN (WGAN)~\cite{arjovsky2017wasserstein} and WGAN-GP~\cite{salimans2016improved}. As reported in~\cite{zhuang2020adabelief}, using each optimizer, we train the model for 100 epochs, generating 64,000 fake images from noise. We compute the Frechet Inception Distance (FID) score between the fake images and the real dataset to assess the generative models. Table~\ref{table:gan-fid} reports the final FID score (lower is better). AdaPlus gets the third-lowest FID score when training WGAN and achieves the lowest FID score when training WGAN-GP. In particular, AdaPlus again outperforms AdaBelief, which demonstrates that aside from precise stepsize adjustment, simultaneously integrating Nesterov momentum and decoupled weight decay helps boost the stability when training GANs.


\section{Related Work}

Unlike SGDM~\cite{sutskever2013importance}, adaptive methods dynamically scale the gradient according to the EMA of the past gradients. Representative adaptive methods include AdaGrad~\cite{duchi2011adaptive},  RMSprop~\cite{tieleman2012lecture}, and Adam~\cite{kingma2014adam}, which enjoy fast speed in the early training period yet exhibit poorer generalization ability than SGDM. Apart from Nadam~\cite{dozat2016incorporating}, AdamW~\cite{loshchilov2017decoupled}, and AdaBelief~\cite{zhuang2020adabelief}, other variants of Adam also have been proposed (e.g., Yogi~\cite{zaheer2018adaptive},  RAdam~\cite{liu2019variance}, AMSGrad~\cite{reddi2019convergence},  AdaMomentum~\cite{wang2021rethinking}, and Adan~\cite{xie2022adan}). These adaptive methods target to achieve the same goal\textemdash accelerating the training and improving the generalization at the same time. Very recently, The XGrad~\cite{guan2023xgrad} framework was proposed which incorporates weight prediction~\cite{guan2019xpipe} into the DNN training to boost the convergence and generalization of gradient-based optimizers.

\section{Conclusions}
\label{sec:majhead}

This paper proposes a novel and efficient adaptive method AdaPlus which combines the benefits of AdamW, Nadam, and AdaBelief and does not introduce any extra parameters.  The extensive experiment evaluations demonstrate that AdaPlus outperforms the other eight state-of-the-art optimizers in terms of simultaneously considering convergence trait, generalization ability, and training stability.


\bibliographystyle{IEEEbib}
\bibliography{strings,refs}

\end{document}